# Multidimensional Digital Smoothing Filters for Target Detection

*Hugh L. Kennedy*

Defence and Systems Institute, University of South Australia

hugh.kennedy@unisa.edu.au

**ABSTRACT**

Recursive, causal and non-causal, multidimensional digital filters, with infinite impulse responses and maximally flat magnitude and delay responses in the low-frequency region, are designed to negate correlated clutter and interference in the 'background' and to accumulate power due to dim targets in the 'foreground' of a surveillance sensor. Expressions relating mean impulse-response duration, frequency selectivity and group delay, to low-order linear-difference-equation coefficients are derived using discrete Laguerre polynomials and discounted least-squares regression, then verified through simulation.

## 1. INTRODUCTION

Low-pass digital filters, such as those proposed by Savitzky and Golay, with a "maximally flat" magnitude and delay response, have smoothing properties in the time domain. This 'duality' makes it possible to derive their filter coefficients in either the frequency *or* time domains [1]-[4]. For instance, Savitzky-Golay filters [5], may be derived by least-squares fitting a polynomial (of degree $B$) to a sampled input sequence over a finite sliding window to yield low-pass filters with a finite impulse response (FIR). The fitted polynomial resulting from this *analysis* process, is evaluated at the center of the odd analysis window to yield linear-phase (smoothing) filters; evaluation between samples yields fractional-delay (interpolating) filters, evaluation at more recent non-central samples yields filters with a reduced group delay; whereas evaluation at future samples yields predictive (extrapolating) filters. The offset ($q$) chosen for the evaluation, or *synthesis*, therefore determines the phase response of the filter [6]. Savitzky-Golay differentiators [7], are obtained by differentiating the fitted polynomial prior to evaluation. FIR

Savitzky-Golay filters are realized using either non-recursive or recursive structures; however care is required in the latter case to avoid rounding error accumulation due to pole-zero cancellation on the unit circle [6].

'Fading-memory' variants of these 'finite-memory' Savitzky-Golay filters may similarly be derived by performing a least-squares fit with an exponentially-decaying error-weighting function (whose $\mathcal{Z}$ transform has a pole at $z = p$, where $p = e^{\sigma}$), yielding recursive structures with an infinite impulse response (IIR) and with stability guaranteed (for all $q$, if $|p| < 1$) [8]-[11]. They are commonly used in target tracking systems to overcome problems of divergence experienced by Kalman filters in the presence of model mismatch [9]; however in this context, they are usually restricted to applications where the time interval between target detections is constant and where data association is unambiguous; furthermore, startup transients must be handled properly. These restrictions have recently been addressed in [12]; with an expanding-memory filter used during track establishment and a fading-memory filter used thereafter; measurements are probabilistically weighted, however the revisit interval is assumed constant.

For satisfactory tracking performance at a reasonable computational cost, these types of 'detect-before-track' systems require the signal-to-noise ratio (SNR) to be relatively large so that: 1) the density of false-detections – due to clutter, interference and noise – is low; 2) the probability of target detection is high; and 3) the measurement error is low. When these conditions cannot be met, methods that exploit spatiotemporal energy distributions in the underlying sensor 'image', i.e. 'track-before-detect' methods are more appropriate [13]-[17]. Target confirmation decisions in both detect-before-track and track-before-detect frameworks are usually framed as hypothesis tests, based on a test statistic involving a likelihood ratio. The likelihood functions for the true and false detections generally have simple idealized forms – Gaussian for the target and Poisson for the clutter/noise, in spatial coordinates. The Rician distribution is ideal for modeling intensity distributions because it results in Rayleigh and Gaussian distributions at the low- and high-SNR limits, respectively [14]. When the assumed forms or the estimated parameters of the underlying distributions are inappropriate or in error, severely degraded target detection and tracking performance results [16].

If not handled properly, structured backgrounds have the potential to 'wreak havoc' in detect-before-track and track-before-detect systems alike. While often ignored in theoretical works, correlated clutter/noise is commonplace in long-range surveillance systems involving infrared search and track cameras, high-frequency radar, passive (bi-static) radar and sonar [16], especially when very weak targets are sought.

The use of fading-memory target-tracking filters, derived using discounted least-squares and orthogonal polynomials, has been thoroughly explored in the literature, in a *detect-before-track* context [8]-[12]; indeed, it is interesting and illuminating that the same filter coefficients are reached from such a variety of different starting points. However, these filters are arguably a more natural solution to the target-tracking problem in a *track-before-detect* context, where uniform sampling rates are guaranteed in the spatial domain (i.e. within a frame or dwell) and expected in the temporal domain, under normal operating conditions (although there may be some "jitter"). Furthermore, the complication of data association is avoided and special logic need not be implemented to handle initialization and start-up transients, although this may be required in the spatial dimensions if a reduction in the sensor's field of regard is unacceptable. In this paper, multidimensional forms of the filters are used to perform both background whitening *and* target enhancement functions.

Strictly speaking, the proposed algorithm is neither a detect-before-track nor a track-before-detect approach because no attempt is made to establish, and maintain continuity, of target identity. As a consequence, data association is avoided, thus the computational load is constant and data independent, i.e. it does not depend on the density or intensity of the target or clutter; furthermore, the filters are amenable to parallelization because the same operations are applied to every 'cell'. The resulting SNR enhancement should improve the performance and simplify the structure of any 'downstream' detect-before-track stage that follows. The proposed algorithm might therefore be regarded as an 'enhance-before-detect' approach. Non causal (forward/backward) IIR filters are used in the spatial dimensions, whereas an IIR filter with a tunable group delay, is used in the temporal dimension. The simple premise underlying the derivation of the filters allows them to be intuitively adapted and tuned for a wide range of functions.

Analog filter prototypes are used to design the multidimensional IIR filters in [18],[19]; whereas, a direct digital design approach is adopted here. Classical analysis offers the designer an array of well-established relationships to build analogue filters; however they do not transfer exactly into the digital domain so a 'sympathetic' discretization method must be chosen to ensure that the intent of the original design is preserved, which adds an extra layer of complexity to the design process.

Matters relating to the use of multidimensional IIR filters, with maximally-flat responses, in enhance-before-detect algorithms, are addressed in this paper: 1) closed-form expressions for the coefficients of low-order linear-difference-equations in terms of the forgetting factor ($\sigma, p = e^{\sigma}$) and the synthesis offset ($q$) are derived; 2) relationships between these design parameters and the frequency response (magnitude and phase) of the filter are described; 3) ways in which the filter response influences the performance of the enhance-before-detect algorithm are discussed; and 4) a technique for estimating point-target velocity by exploiting the local "Laguerre spectrum" is proposed. A particular filter arrangement that is very well-suited to enhance-before-detect roles is also presented – A background subtraction stage is cascaded with a foreground accumulation stage; both stages use non-causal filters in the spatial dimensions and causal filters in the temporal dimension.

Not all of the relationships required for the task at hand have been tabulated in the literature, for instance, phase control is omitted in [9], only causal filters are considered in [8], and the discussion in [6] is limited to (recursive) FIR filters, with pole-zero cancellation on the unit circle, which is good for efficiency but bad for immunity to rounding error accumulation. Non-causal IIR smoothers and differentiators are presented in [20] however the treatment is restricted to first- and second-order filters. Frequency-domain properties are not analyzed in [8]-[11] and the usefulness of analysis-only operations, to yield the Laguerre spectrum [8], is typically overlooked in the modern literature. Recursive analysis-only filters are also derived and applied in this paper.

There are a number of other non-iterative closed-form techniques for deriving the coefficients of low-pass digital filters with maximally-flat responses, that resemble the much-loved monotonic responses of classical, Bessel and Butterworth, analogue-filters [21]-[27].

In Hermann's early treatment of the problem, exactly linear-phase FIR solutions that satisfy magnitude flatness constraints up to a specified derivative order, at frequencies $\omega = 0$ and $\omega = \pm\pi$, are derived [21]. Low-latency low-pass FIR filters may also be designed to satisfy magnitude and group delay flatness constraints at $\omega = 0$ only, which allows the group delay to be varied [22],[23]. Using the closed-form expressions in [24], specification of the filter order, the desired group delay, and the number of zeros at $z = -1$, yields filters with good phase linearity at low-frequencies and very good high-frequency attenuation; however, the ability to control near-DC gain, i.e. bandwidth and roll-off, is limited for low-order filters. In an extension of this work, "partially flat" FIR filters with derivative-constraints are investigated in [25].

Alternatively, maximally-flat IIR filters may be designed using *continuous Legendre polynomials* in the *frequency domain* [26]. Closed-form expressions for the solutions have been given in terms of the order of the numerator and denominator polynomials comprising the discrete-time transfer function of the filter, along with the desired group delay [27]. This is a very flexible technique that allows a wide range of low-order filters to be designed with good phase and magnitude properties; although unfortunately, it is difficult to know in advance whether a stable causal filter will be produced for a given combination of design parameters; furthermore, the magnitude response of the filter – i.e. whether the high-frequency gain is positive or negative – can be very sensitive to small changes in the desired group delay. These 'quirks' are a consequence of the low-frequency flatness constraints imposed on the IIR solution, which may be difficult to satisfy for some order/delay combinations.

The approach proposed in this paper may also involve a few iterations before the required filter response is attained; however, the link between the design parameters and the achieved filter response is easier to predict and understand, using the well-known properties of linear regression, with *discrete Laguerre polynomials*, directly in the *spatiotemporal domain*.

Closed-form solutions to the causal and non-causal regression problem (for $B = 2$) are given in Section 3, resulting in expressions relating the impulse response and frequency response (phase and magnitude) parameters ($p$ & $q$) to third-order linear-difference-equation coefficients ($\boldsymbol{b}$ & $\boldsymbol{a}$); tuning considerations are also discussed. The rationale behind their derivation is given in Section 2, allowing higher order filters to be derived, if required; implementation notes are given briefly in Section 4; and the link between the response parameters and the ability of these filters to enhance the detectability of a dim point-target in correlated clutter/interference is considered in Sections 5 & 6, using simulation.

## 2. FORMULATION

A background-cancelling filter (stage 1) is cascaded with a foreground-enhancing filter (stage 2). The background-cancelling filter is used to remove non-target related correlation from the sensor data due to clutter or interference. The intensity of the background is estimated (or 'predicted') and *subtracted* from the raw input image, ideally leaving foreground features plus white noise. The output of this estimation-error filter, or prediction-error filter, is then passed to the foreground-enhancing filter which is used to increase the SNR of weak point-like target signals 'buried' in white noise. The intensity of foreground features is estimated and 'integrated' over space and time, i.e. *accumulated* over adjacent pixels and frames. Without the prior pre-whitening stage, this filter would also enhance the background and conceal the target features.

One of the attractive features of the proposed approach is that both stages use the same general linear model and have the same general filter structure – The background and foreground features are assumed to be well represented by a local multidimensional polynomial model in the spatial and temporal dimensions. A polynomial of low degree is sufficient to capture 'rolling' low-frequency 'undulation' in the background due to clutter/interference and the local curvature and motion of point-like target returns in the foreground with a (possibly non-isotropic) Gaussian point-spread function (PSF). The use of a Gaussian spatial impulse response would be ideal; however, recursive, low-order linear-difference-equations are not easy to derive for these filters [20],[28]. Three-dimensional (3-D) filters are used here to fully utilize spatiotemporal correlation due to foreground motion and background translation. They are implemented efficiently as (separable and recursive) IIR filters in each dimension. Both stages employ non-causal filters in the spatial dimensions and a causal filter, with a tunable group-delay, in the temporal dimension.

In summary, the filters used by both stages rely on local polynomial models to estimate the background or foreground signal. The stages differ in the way that the fitted polynomials are used – The first stage is a high-pass background-subtracting filter; the second stage is a low-pass foreground-accumulating filter.

Both stages assume that the input has the following form:

$$I(n_x - m_x, n_y - m_y, n_z - m_z) = \sum_{k_x=0}^{B} \sum_{k_y=0}^{B} \sum_{k_z=0}^{B} \beta_{k_x k_y k_z}(n_x, n_y, n_z)\psi_{k_x}(m_x)\psi_{k_y}(m_y)\psi_{k_z}(m_z) \quad (1a)$$

$$J(n_x, n_y, n_z) = I(n_x, n_y, n_z) + \varepsilon \quad (1b)$$

where:

- $I$ is the "noise-free" 3-D signal and $J$ is the "noise-corrupted" 3-D image, as measured by the sensor;
- $x$ and $y$ refer to the spatial dimensions (sampled to yield 'pixels') and $z$ refers to the temporal dimension (sampled to yield 'frames');
- $n$ and $m$ are the global cell (or voxel) and local offset indices, respectively, with indexes increasing in opposing directions, in accordance with filtering convention;
- $\varepsilon$ is a Gaussian-distributed sensor noise term, with $\varepsilon \sim \mathcal{N}(0, \sigma_\varepsilon^2)$;
- $\beta$ are the (cell-dependent) model coefficients;
- $B$ is the model degree, assumed to be the same in each dimension, for convenience; and
- $\psi_k$ is the $k$th (real) local basis function.

The basis functions are constructed by orthonormalizing polynomial components in each dimension, using a linear combination

$$\psi_k(m) = \sum_{\acute{k}=0}^{B} \alpha_{k\acute{k}} m^{\acute{k}} \quad (2)$$

where the $\alpha$ coefficients are determined using the Gram-Schmidt procedure such that

$$\sum_{m=-\infty}^{+\infty} \psi_{k_2}(m) w_\pm(m) \psi_{k_1}(m) = \delta_{k_1 k_2} \quad (3a)$$

in the $x$ and $y$ dimensions and

$$\sum_{m=0}^{\infty} \psi_{k_2}(m) w_+(m) \psi_{k_1}(m) = \delta_{k_1 k_2} \quad (3b)$$

in the $z$ dimension, where $\delta$ is the Kronecker delta function and $w(m)$ is a (non-normalized) weighting function with

$$w_\pm(m) = e^{\sigma|m|} \quad (4a)$$

in the $x$ and $y$ dimensions and

$$w_+(m) = e^{\sigma m} \quad (4b)$$

in the $z$ dimension. Note that numerical orthonormalization may be problematic/inaccurate for some combinations of $\sigma$ and $B$, if the polynomial components are too similar over the specified scale. The $\sigma$ parameter (with $\sigma$<0, so that the resulting filter poles are inside the unit circle) is a 'forgetting' factor, which is assumed to be the same in each dimension, in this Section, for notational simplicity. The infinite summations required to form the orthonormal basis may be found by taking the (one-sided) $\mathcal{Z}$ transform of the summation argument products, multiplying by the $\mathcal{Z}$ transform of the accumulation operation $z/[z-1]$, and invoking the final-value theorem, which simplifies to

$$\mathcal{Z}\{m^{\acute{k}_2}w_+(m)m^{\acute{k}_1}\}\big|_{z=1} + \mathcal{Z}\{(-m)^{\acute{k}_2}w_+(m)(-m)^{\acute{k}_1}\}\big|_{z=1} - [m^{\acute{k}_2}w_+(m)m^{\acute{k}_1}]\big|_{m=0} \quad (5a)$$

in the $x$ and $y$ dimensions and

$$\mathcal{Z}\{m^{\acute{k}_2}w_+(m)m^{\acute{k}_1}\}\big|_{z=1} \quad (5b)$$

in the $z$ dimension. Discrete Laguerre polynomials are produced in the one-sided (temporal) case [8]. Note that the 'script' variable $\mathcal{z}$, is used to refer to the complex frequency-domain plane, while the 'regular' variable $z$, is used to refer to the temporal dimension. The variable $k$ is the orthogonal basis-function index; whereas the accented variable $\acute{k}$, is the polynomial component index. Note also that the final term in (5a) is required to remove an $m = 0$ term from the infinite summation, which would otherwise be counted twice in the first two terms.

To assist the theoretical development in this section, the data are assumed to be of infinite extent in both spatial dimensions and all pixels in a given frame are assumed to arrive at the processor from the sensor simultaneously; furthermore, to simplify notation, sample indices and the summations over those indices are abbreviated below using

$$(\boldsymbol{n}) \equiv (n_x, n_y, n_z),\ (\boldsymbol{q}) \equiv (q_x, q_y, q_z),\ (\boldsymbol{m}) \equiv (m_x, m_y, m_z) \text{ and } \sum_{\boldsymbol{m}} \equiv \sum_{m_x=-\infty}^{+\infty} \sum_{m_y=-\infty}^{+\infty} \sum_{m_z=0}^{+\infty} . \quad (6)$$

Similar notation is also used for the basis functions i.e.

$$(\boldsymbol{k}) \equiv (k_x, k_y, k_z) \text{ and } \sum_{\boldsymbol{k}} \equiv \sum_{k_x=0}^{B} \sum_{k_y=0}^{B} \sum_{k_z=0}^{B} . \quad (7)$$

Subscripts may be dropped for brevity when a given relationship or variable applies to more than one dimension.

The maximum likelihood estimates $\hat{\beta}$, of the model coefficients $\beta$, are determined in the usual way, by minimizing the (weighted) sum-of-squared errors (SSE),

$$\mathrm{SSE}(\boldsymbol{n}) = \sum_{\boldsymbol{m}} \epsilon(\boldsymbol{n}-\boldsymbol{m}) w(\boldsymbol{m}) \epsilon(\boldsymbol{n}-\boldsymbol{m}) \quad (8)$$

where

$$\epsilon(\boldsymbol{n}-\boldsymbol{m}) = J(\boldsymbol{n}-\boldsymbol{m}) - \hat{I}(\boldsymbol{n}-\boldsymbol{m}) \quad (9)$$

and

$$\hat{I}(\boldsymbol{n}-\boldsymbol{m}) = \sum_{\boldsymbol{k}} \hat{\beta}_{\boldsymbol{k}}(\boldsymbol{n})\, \psi_{\boldsymbol{k}}(\boldsymbol{m}). \quad (10)$$

In (8) the weighting function is

$$w(\boldsymbol{m}) = w_\pm(m_x) w_\pm(m_y) w_+(m_z) . \quad (11)$$

For an orthonormal basis set, the parameter estimates are simply determined using the following *analysis* equations:

$\hat{\beta}_k(n) = \sum_{m=-\infty}^{+\infty} \psi_k(m) w_\pm(m) J(n-m)$ (12a)

in the $x$ and $y$ dimensions and

$\hat{\beta}_k(n) = \sum_{m=0}^{+\infty} \psi_k(m) w_+(m) J(n-m)$ (12b)

in the $z$ dimension. Taking the $\mathcal{Z}$ transforms of the $\psi_k(m)w(m)$ products allows the convolutions to be evaluated recursively. With the model parameters $\hat{\beta}_{\boldsymbol{k}}(\boldsymbol{n})$, or the so-called Laguerre spectrum, in the vicinity of $\boldsymbol{n}$ determined using (12), the 'noise-free' estimate of the input sequence may then be evaluated at $\boldsymbol{n} - \boldsymbol{q}$, using the *synthesis* equation:

$\hat{I}(\boldsymbol{n} - \boldsymbol{q}) = \sum_{\boldsymbol{k}} \hat{\beta}_{\boldsymbol{k}}(\boldsymbol{n}) \psi_{\boldsymbol{k}}(\boldsymbol{q})$ (13)

which is then used to form the estimation residual at a specific cell

$I_\epsilon(\boldsymbol{n} - \boldsymbol{q}) = J(\boldsymbol{n} - \boldsymbol{q}) - \hat{I}(\boldsymbol{n} - \boldsymbol{q})$. (14)

In general, the elements of $\boldsymbol{q}$ need not be integers in (13), however integer values must be used if (14) is to be used. Substituting the $\mathcal{Z}$ transforms of (12a) or (12b) into (13) yields a linear difference equation (LDE) for a low-pass IIR filter. For the non-causal (forward/backward) filters, $q_x = q_y = 0$; for causal filters, $I_\epsilon$ is formed by delaying $\hat{I}$ by $|q_z|$ frames if $q_z < 0$ (predictive case), whereas $J$ is delayed by $q_z$ frames if $q_z > 0$. Using $B = 1$ with $q_z < 0$ and $q_z > 0$ in the temporal dimension yields the lead and lag filters in [29], respectively. In the current application, when (14) is evaluated at $\boldsymbol{q}$ over all pixels in the current frame, $I_\epsilon$ is the whitened image, with the background removed, which is output by stage 1. The operation in (14) converts the low-pass filtered output $\hat{I}$, into a high-pass filtered output $I_\epsilon$.

In some applications it is desirable to evaluate the weighted SSE at every pixel, to give an indication of how well the local polynomial model fits the input data around the cell at $\boldsymbol{n}$, using

$\mathrm{SSE}(\boldsymbol{n}) = \sum_{\boldsymbol{m}} J(\boldsymbol{n} - \boldsymbol{m}) w(\boldsymbol{m}) J(\boldsymbol{n} - \boldsymbol{m}) - 2\sum_{\boldsymbol{m}} J(\boldsymbol{n} - \boldsymbol{m}) w(\boldsymbol{m}) \hat{I}(\boldsymbol{n} - \boldsymbol{m}) + \sum_{\boldsymbol{m}} \hat{I}(\boldsymbol{n} - \boldsymbol{m}) w(\boldsymbol{m}) \hat{I}(\boldsymbol{n} - \boldsymbol{m})$ (15)

which may also be evaluated recursively by taking $\mathcal{Z}$ transforms. In the current application, the second-stage is a simple quadratic filter that accumulates the weighted power of the 3-D polynomial fitted to the target signal using only the last term of (15) i.e.

$\hat{P}(\boldsymbol{n}) = \sum_{\boldsymbol{m}} \hat{I}(\boldsymbol{n} - \boldsymbol{m}) w(\boldsymbol{m}) \hat{I}(\boldsymbol{n} - \boldsymbol{m})$. (16)

Substitution of (10) into (16) yields

$\hat{P}(\boldsymbol{n}) = \sum_{\boldsymbol{m}} \left[\sum_{\boldsymbol{k}_2} \hat{\beta}_{\boldsymbol{k}_2}(\boldsymbol{n}) \psi_{\boldsymbol{k}_2}(\boldsymbol{m})\right] w(\boldsymbol{m}) \left[\sum_{\boldsymbol{k}_1} \hat{\beta}_{\boldsymbol{k}_1}(\boldsymbol{n}) \psi_{\boldsymbol{k}_1}(\boldsymbol{m})\right]$ (17)

which, due to basis set orthonormality, simplifies to

$\hat{P}(\boldsymbol{n}) = \sum_{\boldsymbol{k}} \hat{\beta}_{\boldsymbol{k}}^2(\boldsymbol{n})$ (18a)

or

$\hat{P}(\boldsymbol{n}) = \sum_{\boldsymbol{k}\in\boldsymbol{\Omega}} \hat{\beta}_{\boldsymbol{k}}^{2}(\boldsymbol{n})$ (18b)

if it is known *a priori* that targets of interest only have non-negligible power in a subset ($\boldsymbol{\Omega}$) of bins (see Appendix B). A weighted sum of coefficients could instead be used in (18) to favor targets with specific shapes and velocities [30]; however this slows down execution by adding another loop over all possible target classes. As the weighting functions used here are not normalized, the average signal power $\hat{P}_{\text{avg}}(\boldsymbol{n})$ in the vicinity of $\boldsymbol{n}$ is determined using $\hat{P}_{\text{avg}}(\boldsymbol{n}) = c_{\text{nrm}}\hat{P}(\boldsymbol{n})$, where the normalizing factor is determined using (5), i.e.

$c_{\text{nrm}} = \{[2/(1-p_x)-1][2/(1-p_y)-1][1/(1-p_z)]\}^{-1}.$ (18c)

## 3. PARAMETERIZATION

Responses of a variety of filter tunings are given in Figs. 1-3. Both causal and non-causal variants are shown in Fig. 1. In all cases $B = 2$, i.e. a quadratic model. These types of low-pass filters are used to generate $\hat{I}$ in stage 1, i.e. background estimation. As $\sigma$ becomes less negative, the (real & repeated) filter poles at $p = e^{\sigma}$ move radially outward from the origin of the $z$ plane towards the unit circle, which increases frequency selectivity and decreases temporal/spatial selectivity. For filters with a near-unity pole radius, frequency selectivity further increases as $q$ becomes more positive, i.e. as the nominal group-delay increases. The 'nominal' qualifier is used here to highlight the fact that the phase response is only approximately linear in the low-frequency pass-band. Note that $\sigma$ (or $p$) determines the pole radius and influences the zero locations; whereas $q$ only influences the zero locations, in the analysis & synthesis filters (only used in stage 1). By definition, the non-causal filters have a group-delay of zero, and as expected, greater frequency selectivity (i.e. flatter low-frequency gain and greater high-frequency attenuation) than their zero-latency ($q = 0$) causal counterparts. This is due to the symmetry (or anti-symmetry) of their impulse responses around the 'current' sample (at $m = 0$).

The implementation described in Section 4, which was used to process the data described in Section 5, used $\sigma_x = \sigma_y = -1/2$, $q_x = q_y = 0$, & $B = 2$ in stages 1 and 2; $\sigma_z = -1/4$ and $q_z = 4$ was used in stage 1, whereas $\sigma_z = -1/2$ was used in stage 2. The filter coefficients ($\boldsymbol{b}$ & $\boldsymbol{a}$) derived using these parameters are used in the (causal and recursive) linear-difference-equation

$\sum_{m=0}^{M_a} a_m y(n-m) = \sum_{m=0}^{M_b} b_m x(n-m)$ (19a)

or

$y(n) = \sum_{m=0}^{M_b} b_m x(n-m) - \sum_{m=1}^{M_a} a_m y(n-m)$ (19b)

for $a_0 = 1$, where $y(n)$ and $x(n)$ are the $n$th filter outputs and inputs, respectively. This equation implements the following discrete-time transfer function, which has an IIR and a maximum order of 3 ($M_a = 3$), for $B = 2$:

$$H(z) = \frac{b_0 + b_1 z^{-1} + b_2 z^{-2} + b_3 z^{-3}}{a_0 + b_1 z^{-1} + a_2 z^{-2} + a_3 z^{-3}}. \quad (20)$$

The low-pass analysis & synthesis filters that produce $\hat{I}(\boldsymbol{n})$ in stage 1 are given in Table I; the analysis-only filters that produce $\hat{\beta}_{\boldsymbol{k}}(\boldsymbol{n})$ in stage 2 are given in Table II. Non-causal filters are realized by summing the outputs of separate filters applied to the input data in the forward (fwd) and backward (bwd) directions. For symmetric basis components (e.g. $k = 0$ & $k = 2$), the filter coefficients are the same; for anti-symmetric components (e.g. $k = 1$), some coefficients change sign. Care is required when interpreting the output of the analysis-only filters, because $m$ and $n$ increase in opposite directions.

The frequency response $H(\omega)$, of the filter is found by substituting $z = e^{j\omega}$ into (20). Using the stage 1 filter coefficients given in Table I in (20) and evaluating derivatives of $|H(\omega)|^2$ at $\omega = 0$, reveals that the first, second and third derivatives are all equal to zero, confirming that the procedure does indeed result in some degree of flatness.

The relationships used to determine the coefficient values in Tables I & II are given in Table III. A worked example presented in Appendix A, illustrates how the expressions in Table III are derived using (1) to (13). Only closed-form expressions for third-order filters resulting from quadratic polynomial models ($B = 2$) have been derived, as these yield an optimal balance between versatility and simplicity. For $q = 0$ the expressions for the (zero-latency) causal filters used in stage 1 are the same as the third-order range-filter in [9] and the position element of the state-space recursive least-squares estimator in [10]; the causal stage 1 coefficients, for any $q$, may also be derived using (13.3.11) in [8]. When realized in canonical form, with 6 multipliers and 5 adders, the filter is less complex than the corresponding recursive FIR implementation in [6], which uses 5 multipliers and 12 adders. (Note, the author believes that the second term of (11) in [6] should have its sign reversed.)

## 4. IMPLEMENTATION

The proposed enhance-before-detect algorithm was implemented on a personal computer with a 64 bit operating system and an Intel ® i7-4810MQ CPU @ 2.8GHz using an interpreted MATLAB ® script. The processing architecture is summarized as follows: $J \xrightarrow{\text{stage 1}} I_\epsilon \xrightarrow{\text{stage 2}} \hat{P}_\epsilon$. Thus stage 1 involves analysis and synthesis operations, which are combined to form a single separable IIR filter in each dimension, as described in (1) to (14); whereas stage 2 involves only analysis operations, as described in (1) to (12), which are implemented as a bank of IIR filters in each dimension, followed by the summation in (18).

Stage 2 is therefore somewhat more expensive than stage 1 because its software implementation involves nested loops over $k_x$, $k_y$ and $k_z$ to populate the 3-D array containing the $\hat{\beta}_{k_x k_y k_z}$ coefficients; however, some coefficient combinations may be omitted for some target classes (see Appendix B). As the model coefficients are pixel dependent, a 5-D array is required to store the estimates in the current frame. Both stages avoid loops over $n_x$ and $n_y$ by using vectorized MATLAB operations. Loops over $\boldsymbol{m}$ are of course avoided through the use of recursion. In both stages, the MATLAB `filter()` command was used to process the rows and columns of each frame. In a given spatial dimension, the non-causal IIR filters were realized by processing the input data once in the forward direction, then again in the backward direction by 'flipping' the input data in either the up/down or left/right directions. The causal IIR filters in the temporal dimension were implemented using delayed frame buffers for the raw input and processed output.

At an operational level, the computational cost is dominated by the number of multiply and accumulate operations (MACs) associated with the spatial and temporal IIR filters employed in stages 1 and 2; there are also a few accompanying addition and subtraction operations which are not counted in the following analysis. Inspection of the non-zero and non-unity filter coefficients in Tables IA, IB, IIA & IIB gives an *approximate* indication of the computational complexity of the proposed algorithm, per pixel per frame. There are $4 \times 7 + 6$ MACs associated with the application of the stage 1 filters. The factor of four is due to application of the non-causal spatial filters in the left, right, upward and downward directions. If all bins of the Laguerre spectrum are evaluated, there are $(2 \times 13) + 3(2 \times 13) + 9(12)$ MACs associated with application of the stage 2 filters in the $x$, $y$ and $z$ dimensions, followed by $3 \times 3 \times 3$ MACs to implement (18a). For frames with dimensions $128 \times 128$, with $B = 2$ in all dimensions in both stages and when only 7 Laguerre spectrum bins are evaluated in stage 2 – as per (18b) and as explained in Appendix B – a throughput rate of approximately 88 frames per second was achieved.

## 5. SIMULATION

The performance of the filters was analyzed using simulated, (pseudo-) randomly-generated input-data, which were intended to be a 'crude caricature' of a detection/acquisition scenario in either an airborne infrared camera (long-range target set against blue-sky), or a sky-wave radar (small target at night, with Doppler information removed). A total of 64 frames of $128 \times 128$ pixels were generated per scenario. A (real-valued) background texture was synthesized using 10 randomly generated sinusoidal components with: a spatial frequency randomly distributed over an interval of 0 to 1/33 cycles per pixel, a random phase, and an amplitude of 1/10. The clutter/interference texture was translated using a randomly generated (group) velocity of $\boldsymbol{v}_{\mathrm{clt}}$. The components of $\boldsymbol{v}_{\mathrm{clt}}$ were uniformly distributed between 0 and 1 (pixels/frame). An additional DC component with an amplitude of 1 was also included in the background. A foreground target was then *added* with: a maximum intensity of 1, a Gaussian PSF with a standard deviation (std. dev.) of 2 pixels and a random velocity of $\boldsymbol{v}_{\mathrm{tgt}}$. The components of $\boldsymbol{v}_{\mathrm{tgt}}$ were uniformly distributed between -1 and 0 (pixels/frame). The target's position in the final frame is $\boldsymbol{p}_{\mathrm{tgt}} = [64,64]$ (pixels), displaced by a random offset $\Delta\boldsymbol{p}_{\mathrm{tgt}}$, of 0 to 1 pixels. Finally, white Gaussian noise was added to each frame with a std. dev. of $\sigma_{\varepsilon}$. In the example shown in Fig. 4, $\boldsymbol{v}_{\mathrm{clt}} = [0.51,0.46]$, $\boldsymbol{v}_{\mathrm{tgt}} = [-0.73,-0.07]$ and $\Delta\boldsymbol{p}_{\mathrm{tgt}} = [0.92, 0.54]$. The sensor noise was generated to yield a (target) SNR of 6 dB ($\sigma_{\varepsilon} \cong 0.5$).

## 6. DISCUSSION

### *6.1 Stage 1 Analysis*

Reducing $B$ in stage 1 resulted in a slight decrease in the SNR enhancement and a slight increase in the frame rate. This is because decreasing $B$ decreases the filter order and the -3 dB filter bandwidth, of the low-pass filters. For the randomly selected scenario depicted in Fig. 4 (i.e. the "selected scenario"), the average approximate SNR of the stage 2 output for $B = 2, 1$ & $0$, in stage 1, is 12.4, 11.7 & 11.8 dB, respectively. The discrete-time transfer function of the background subtraction filters used in stage 1 is required to account for the reasonable performance observed for the reduced-degree models. Each low-pass filter (lpf), designed using a delay of $q$, is converted into a high-pass filter (hpf) using (14); taking the $\mathcal{Z}$ transform yields

$$H_{\mathrm{hpf}}(z) = z^{-q} - H_{\mathrm{lpf}}(z). \tag{21}$$

The frequency response of the resulting hpf is shown in Fig. 5 for a non-causal filter ($q = 0$) and causal filters with $q = 0$ & $4$. The responses for the lpf prototypes presented in Section 3 are quite good with magnitude and phase approximately linear in the low-frequency region; however this is not carried over into the hpf version, because the resulting high-frequency pass-band is wider and not so well defined. Using a quadratic model ($B = 2$), for a lpf pole multiplicity of 3, increases the width of the DC notch in the hpf, so that the background spatial components used in the simulation (with $f \leq 0.03$) all fall well within the notch of the non-causal spatial filters. Increasing the latency of the causal filter using $q = 4$ instead of $q = 0$ decreases the high-frequency attenuation which allows more of the target signal to reach stage 2. Using a simple two-sided moving exponential average ($B = 0$) as the lpf prototype in the spatial dimensions does still give reasonable attenuation near DC and near unity gain (0 dB) away from DC. The discrete-time transfer function for this simple hpf is derived using (21) and a smoothing lpf prototype, which may be derived using the procedure described in Section 2; alternatively, the hpf may be adapted from [20]:

$$H_{\mathrm{hpf}}(z) = 1 - \frac{1-p}{1+p}\left(\frac{1}{1-p/z} + \frac{1}{1-pz} - 1\right). \quad (22)$$

The proposed design technique produces filters with repeated poles on the real axis ($0 < p < 1$), thus the ability to widen the pass band of the lpf is limited; unfortunately, increasing $B$ has diminishing returns. One of the benefits of the method described in [25], is the ability to generate solutions with complex poles, which allows low-pass filters with high cut-off frequencies to be realized with low-order filters.

Translation of the spatial components due to apparent background motion causes the spatial spectrum to be 'tilted' out of the $x$-$y$ plane in the 3-D frequency domain, where $f_z = 0$, according to

$$f_z = -v_x f_x - v_y f_y. \quad (23)$$

This relationship [31], may be used to determine the width of the clutter/interference notch required of the temporal filter used in stage 1. Using $f_x = f_y = \pm 0.03$ and $v_x = v_y = \pm 0.5$ in (23) yields $f_z = \pm 0.03$. The spatial filters used in stage 1 (causal, $B = 2$, $\sigma = -1/4$ & $q = 4$) apply an attenuation of 20 dB at this frequency but only 6 dB if the speed is doubled (see Fig 5); therefore, degraded target enhancement is expected due to residual background clutter/interference in the stage 1 output. Applying factors of 1, 2 & 4x to the velocity components of the background in the selected scenario results in average output SNRs of 12.4, 9.5 & 7.7 dB, respectively.

As $\sigma \to 0$ from the left (i.e. as $p \to 1$ from the left, where $p = e^{\sigma}$) the frequency selectivity (i.e. pass-band flatness and stop-band attenuation) of the background estimation filter in stage 1 increases in the spatial dimensions but the width of transient phenomena around the perimeter of the processed image also increase due to the longer mean impulse response. This has the effect of reducing the effective coverage of the sensor. Longer (mean) impulse responses are more useful in the temporal dimension, where startup transients are quickly 'forgotten', especially if the parameters of the background change only slowly (i.e. approximately wide-sense stationary). When $\sigma$ is doubled, for a shorter average impulse response, in all dimensions in stage 1, the average output SNR in the selected scenario decreases by 5.4 dB.

Using $q_z = 4$ (i.e. a delay of four frames) in stage 1 also promotes frequency selectivity (see Fig. 2), to ensure that the low-frequency background is strongly attenuated in $I_{\epsilon}$, while preserving most of the high-frequency foreground content (see Fig. 5). However, using $q_z \leq 0$ may be beneficial in closed-loop control/guidance systems, to minimize delays and maximize stability margins [29], at the expense of frequency selectivity. To maximize noise attenuation, $q_z$ should be chosen to suit the selected value of $p$ in the $z$ dimension (i.e. $p_z$). If the spatiotemporal clutter frequencies are within the passband of the prediction filter employed in stage 1, i.e. if the selected polynomial degree ($B$) is a reasonable approximation, then the so-called variance reduction factor (VRF), is a useful indication of the expected sensor noise attenuation in stage 1. As defined in [8] &

[32], the VRF is the ratio of the signal estimate variance – i.e. $\mathrm{E}\langle(\hat{I}-I)^2\rangle$, where $\mathrm{E}\langle\cdot\rangle$ is the expectation operator – to the sensor noise variance $\sigma_\varepsilon^2$, where $\sigma_\varepsilon^2 = \mathrm{E}\langle(J-I)^2\rangle$, in the absence of a foreground signal. Using (13.6.20) in [8], the VRF of the temporal causal filter with $B_z = 2$, as a function of $p_z$ and $q_z$ (subscripts dropped below) is

$$\mathrm{VRF} = FAF^{\mathrm{T}} \tag{24a}$$

where

$$F = \left[1 \quad \{p - q + pq\} \quad \left\{2pq(p-1) + \tfrac{1}{2}q(p-1)^2(q-1) + p^2\right\}\right] \tag{24b}$$

and

$$A = (1-p)\begin{bmatrix} \frac{1}{(1+p)} & \frac{1}{(1+p)^2} & \frac{1}{(1+p)^3} \\ \frac{1}{(1+p)^2} & \frac{2}{(1+p)^3} & \frac{3}{(1+p)^4} \\ \frac{1}{(1+p)^3} & \frac{3}{(1+p)^4} & \frac{6}{(1+p)^5} \end{bmatrix}. \tag{24c}$$

Substituting the values used in the $z$ dimension of stage 1 ($p = e^{-1/4}$ and $q = 4$) into (24) and evaluating, yields VRF $\cong 0.1$, which suggests an improvement of 10 dB due to the temporal filtering alone. Taking the derivative of the VRF with respect to $q$, setting it to zero, then solving for $q$, yields the 'optimal' value of $q$, for a given $p$, that minimizes the VRF. For $B = 2$,

$$q_{\mathrm{opt}} = \left[4p - \sqrt{2(p^2 + 4p + 1)} + 2\right]/[2(1-p)]\ . \tag{25}$$

Use of $q_{\mathrm{opt}}$ in the causal low-pass filters in stage 1 (see Table III) places a zero in the complex frequency-domain at $z = -1$, for infinite attenuation at the Nyquist frequency. The VRF decreases, thus the noise attenuation and SNR increase, as $p \to 1$. Substituting the selected value of $p = e^{-1/4}$ into (25) yields $q_{\mathrm{opt}} \cong 4.6$. An integer value is required to form the prediction error in stage 1, as shown in (14), therefore a value of $q_{\mathrm{z}} = 4$ was used. As a rule of thumb, using a value of $q$ close to the centroid of the weighting function $w(m)$ in (4b) generally gives satisfactory results, as might intuitively be expected from the theory of linear regression.

However, minimization of the VRF is not the only consideration when selecting values for the $p$ and $q$ parameters in stage 1. As previously mentioned, decreasing $p$ reduces the impulse response duration, thus the impact of transients, and the associated bias errors, arising from sudden changes in the clutter parameters ($\beta$) in the background; whereas decreasing $q$ reduces the processing latency of the system, which may be a critical consideration in closed-loop systems. The average SNR of the stage 2 output, for the selected scenario is 11.1, 12.4, 12.4 & 11.9 dB, for $q_{\mathrm{z}}$ values of 0, 2, 4 & 6, respectively; however, the expected trend is more noticeable if the structured background is omitted from the simulation, yielding outputs of 11.7, 13.0, 13.6 & 13.4 dB, respectively. These results indicate that the VRF is a reasonable predictor of performance.

Unfortunately, stage 1 is responsible for some target signal attenuation, in the low-frequency region; however, for narrow target PSFs, there is sufficient high-frequency content to stimulate stage 2. The overall performance therefore depends on the overlap between the background and foreground power-density spectra. For instance, very little improvement in target 'visibility' is expected for high-frequency backgrounds and wide target PSFs. Residual structure in the stage 1 output overwhelms the stage 2 accumulator and masks dim targets. In the absence of background clutter/interference, stage 1 could of course be bypassed with $J$ sent directly to stage 2; and when targets are relatively bright, stage 2 may be omitted. When the frequencies of the background components used in Fig. 4 are doubled, the average output SNR decreases by 3.1 dB.

### *6.2 Stage 2 Analysis*

It is interesting to note that using $B = 0$ in all dimensions of stage 2, which is equivalent to simply convolving the output of stage 1 with an exponential PSF that decays exponentially in time, actually improves the detection performance in the selected scenario by approximately 1 dB. Increasing the polynomial degree in all dimensions of stage 2 increases the noise power somewhat, due to the extra degrees of freedom available; however, it also results in a multidimensional *Laguerre spectrum*, as constructed using (12). This process is referred to in the literature as a Discrete Laguerre Transform (DLT) [33]. The resulting spectrum may then be processed to form a feature vector at each pixel in every frame, containing local shape and velocity estimates (i.e. an optical-flow field [34]), for instance. This information may then be exploited by 'downstream' processing stages, e.g. for target classification and/or target tracking purposes, and is the main motivation for using $B = 2$ in stage 2 in this paper. An alternative approach was adopted by the author in [35], where banks of first-order recursive filters with a finite impulse response were used to compute the (short-time/space) *Fourier spectrum* for local motion analysis (using the local 3-D autocorrelation function) and filtering in an infrared target detection and tracking application. In that work, the primary objective was background clutter suppression, as sensor noise was assumed to be negligible, thus further integration of the foreground target signal was not required. A possible technique for exploiting the Laguerre spectrum to generate a velocity estimate is presented in Appendix B. The method is only presented here for illustrative purposes and its properties are not explored in detail. It is simply used demonstrate the feasibility of this approach, and to justify the extra complexity associated with using $B = 2$ in stage 2. It was only applied to the pixel in each frame where the output of stage 2 is maximized. In the selected scenario, the average estimate of the target velocity was $\hat{\boldsymbol{v}}_{\mathrm{tgt}} = [-0.64, -0.08]$ which is not too far from the true value of $\boldsymbol{v}_{\mathrm{tgt}} = [-0.73, -0.07]$. Preliminary investigations indicate that, like the gradient-based Lucas-Kanade method of computing optical flow [34], the velocity estimates tend to be biased (towards zero) and gross errors are expected when speeds start to exceed 1 pixel per frame.

For a perfectly whitened input $I_\epsilon$, stage 2 enhancement performance improves for slower targets with wider PSFs. Fast targets benefit less from target power accumulation in the temporal dimension; however this effect is offset by improved spatial accumulation, if the PSF is broadened (i.e. for near and/or large targets). If no background is added in the selected scenario, and stage 1 is bypassed, the SNR of the stage 2 output is 6.9, 13.0, 19.5 & 21.6 dB, for PSFs of ½, 1, 2 & 4 pixels, respectively. In stage 2, longer impulse responses are required to sufficiently accumulate target signal power when the target SNR is low; however the benefits are only fully realized if the target is slow. For the target speed and PSF combination used in Fig. 4, the target is still sporadically visible in the stage 2 output, with an average output SNR of 8.7 dB, if the input SNR is reduced by 6 dB to 0 dB.

The proposed foreground enhancement technique was compared with a 3-D matched filter. The matched filter is known to be optimal when the target signal is known precisely and when the sensor noise is zero-mean and uncorrelated. The $9 \times 9 \times 9$ convolution kernel of an FIR matched filter was created using the exact velocity and Gaussian PSF of the target and used to process the whitened output of stage 1. This 'clairvoyant' filter was able to attain an impressive output SNR of 16.8 dB in the selected scenario. However, in practice, the velocity of the target is rarely known in advance, therefore a bank of such filters must be applied, each tuned to a possible velocity hypothesis [36],[37]. At each pixel, the filter with the greatest output power is selected. A $3 \times 3$ bank of filters created using $v_x = [-1{,}01]$ and $v_y = [-1{,}01]$ yielded a disappointing output SNR of 10.9 dB and $\hat{\boldsymbol{v}}_{\mathrm{tgt}} = [-1{,}0]$; whereas a $5 \times 5$ bank of filters created using $v_x = \left[-1, -\frac{1}{2}, 0, \frac{1}{2}, 1\right]$ and $v_y = \left[-1, -\frac{1}{2}, 0, \frac{1}{2}, 1\right]$ yielded an improved output SNR of 12.1 dB and $\hat{\boldsymbol{v}}_{\mathrm{tgt}} = [-0.68, -0.01]$, at considerable computational expense, which is still less than the 12.4 dB achieved via the Laguerre spectrum. When the convolution kernels of the matched filters are simply applied (as done here) in the voxel domain, $9 \times 9 \times 9$ MACs are required per velocity hypothesis, which is also substantially more expensive than the proposed method. While frequency domain and/or recursive implementations may reduce the complexity of matched filtering approaches somewhat, a complexity that is proportional to the number of filters in the bank is a burden that is difficult to avoid.

## 7. CONCLUSION

Unlike many other detect-before-track and track-before-detect approaches used to detect barely discernable targets in long-range surveillance-sensor systems, the method proposed in this paper does not assume that targets are the only features in the image that exhibit correlation in space and time. A flexible and efficient multidimensional digital filtering scheme is used to suppress background features and enhance foreground features. Foreground enhancement is achieved using a recursive Discrete Laguerre Transform (DLT); the resulting spectrum may also be used to characterize foreground features using shape and velocity attributes. The IIR filters are manually tuned using a small number of configuration parameters. A direct digital design approach is used to avoid discretization phenomena and optimization/numerical procedures are not required. Like FIR Savitzky-Golay filters, the least-squares formulation in the pixel/frame domain ensures: firstly, that mean-squared errors are minimized, when the foreground or background do indeed adhere to a low-degree polynomial model, over the scales specified by the weighting functions (using the $p$ or $\sigma$ parameters); and secondly, that a maximally-flat low-frequency response is obtained. For causal filters, the trade-off between frequency selectivity and group delay (lag or lead) may be adjusted to suit requirements (using the $q$ parameter). The proposed framework and the closed-form expressions for the filter coefficients are intended to ease the burden of recursive multidimensional filter design.

## 8. APPENDICES

*Appendix A*

The closed form expressions for the LDE coefficients, used in the low-pass (analysis-and-synthesis) filters of stage 1, and the (analysis-only) filters of stage 2, with $B = 2$ (see Table III), are derived in this Appendix, by following the procedure outlined in Section 2. Only the causal case, as used in the temporal dimension, is considered. An analogous approach is taken for the non-causal filters, which for the sake of brevity, is not reproduced here. The first step is to determine the discrete Laguerre polynomial coefficients $\alpha_{k\acute{k}}$ used in (2). This is done via the Gram-Schmidt procedure, which ensures that the resulting polynomials are orthogonal and normalized, as specified in (3b), with respect to the causal weighting function $w_+(m) = e^{\sigma m}$. As indicated in (5b), the infinite summations over $m$ required in this procedure, i.e. $\sum_{m=0}^{\infty} m^{\acute{k}_2} e^{\sigma m} m^{\acute{k}_1}$, are conveniently evaluated via an $m \rightarrow z$ transform using $\mathcal{Z}\{m^{\acute{k}_2} e^{\sigma m} m^{\acute{k}_1}\}\big|_{z=1}$. The resulting coefficients $\alpha_{k\acute{k}}$, for the $k$th orthogonal polynomial $\psi_k(m)$, in terms of the $\acute{k}$th orthogonal polynomial component $\phi_{\acute{k}}(m) = m^{\acute{k}}$, with $\psi_k(m) = \sum_{\acute{k}=0}^{B} \alpha_{k\acute{k}} m^{\acute{k}}$, are given in Table A.I. Note that the discrete Laguerre polynomials used here assume that the weighting function is not multiplied by a normalizing factor.

Convolution with the $k$th exponentially-weighted orthogonal polynomial $w_+(m)\psi_k(m) = e^{\sigma m}\psi_k(m)$, is achieved using a linear combination of the outputs of $B + 1$ component filters. The LDE coefficients of the $\acute{k}$th component filter are derived from the weighted component transfer function $\mathcal{F}_{\acute{k}}(z)$, which is found by taking the $\mathcal{Z}$ transform of the exponentially-weighted polynomial component $w_+(m)\phi_{\acute{k}}(m) = e^{\sigma m}m^{\acute{k}}$, i.e.

$$\mathcal{F}_0(z) = \mathcal{Z}\{e^{\sigma m}\} = \frac{z}{z-p} \quad \text{(A.1a)}$$

$$\mathcal{F}_1(z) = \mathcal{Z}\{e^{\sigma m}m\} = \frac{pz}{(z-p)^2} \quad \text{(A.1b)}$$

$$\mathcal{F}_2(z) = \mathcal{Z}\{e^{\sigma m}m^2\} = \frac{pz(z+p)}{(z-p)^3}\,. \quad \text{(A.1c)}$$

The transfer function of the $k$th (causal) analysis filter $\mathcal{G}_k(z)$, as used in stage 2, and as specified in Table III, is therefore

$$\mathcal{G}_k(z) = \boldsymbol{\alpha}_k\boldsymbol{\mathcal{F}}(z) \quad \text{(A.2)}$$

where $\boldsymbol{\alpha}_k = [\alpha_{k0} \quad \alpha_{k1} \quad \alpha_{k2}]$, i.e. the $k$th row of Table A.I, and $\boldsymbol{\mathcal{F}}(z) = [\mathcal{F}_0(z) \quad \mathcal{F}_1(z) \quad \mathcal{F}_2(z)]^{\mathrm{T}}$, for $B = 2$. As emphasized in Section 2 and as specified in (12b), the output of the $k$th temporal analysis filter $\hat{\beta}_k(n)$, for $n \gg 0$ (i.e. after the start-up transient has effectively passed), is the regression coefficient corresponding to the least-squares fitting of $\psi_k(m)$, to the input sequence $x(n)$, using a weight $w_+(m)$, that decays exponentially with increasing $m$ (and decreasing $n$). In the context of the foreground-accumulating filters employed in stage 2, and in the absence of any spatial filtering, the input is the prediction error $I_\epsilon(n)$. Note that inspection of the $\boldsymbol{\mathcal{F}}(z)$ components in (A.1) suggests that further efficiencies may be obtained using a cascaded filter structure [38]; however, this was not implemented here.

With the regression coefficients $\hat{\beta}_k(n)$ determined using a bank of $B + 1$ parallel analysis filters, a 'noise-free' estimate of the input is 'reconstructed' or 'synthesized' by evaluating the weighted least-squares fit at a point displaced by $q$ samples from the current sample at $m = 0$. The transfer function $\mathcal{H}(z)$ of the resulting causal analysis-and-synthesis filter is

$$\mathcal{H}(z) = \boldsymbol{\psi}(q)^{\mathrm{T}}\boldsymbol{\mathcal{A}}\boldsymbol{\mathcal{F}}(z) \quad \text{(A.3a)}$$

where $\boldsymbol{\psi}(q) = [\psi_0(q) \quad \psi_1(q) \quad \psi_2(q)]^{\mathrm{T}}$ for $B = 2$ and the $k$th row of the square (lower-triangular) matrix $\boldsymbol{\mathcal{A}}$ is equal to $\boldsymbol{\alpha}_k$. Alternatively, using the fact that $\boldsymbol{\psi}(q) = \boldsymbol{\mathcal{A}}\boldsymbol{\phi}(q)$, where $\boldsymbol{\phi}(q) = [\phi_0(q) \quad \phi_1(q) \quad \phi_2(q)]^{\mathrm{T}} = [1 \quad q \quad q^2]^{\mathrm{T}}$

$$\mathcal{H}(z) = \boldsymbol{\phi}(q)^{\mathrm{T}}\boldsymbol{\mathcal{A}}^{\mathrm{T}}\boldsymbol{\mathcal{A}}\boldsymbol{\mathcal{F}}(z). \quad \text{(A.3b)}$$

Expansion and simplification of (A.3b) yields the causal low-pass filter coefficients in Table III, used to estimate the background intensity in stage 1 as defined in (13). In the context of the background-subtraction filters employed in stage 1, and in the absence of any spatial filtering, $\mathcal{H}(z)$ transforms $J(n)$ into $\hat{I}(n)$. This operation is recursively implemented in the time domain at each pixel using (19b).

The gradient of the input at $n - q$ is estimated by determining the derivative of the weighted least-squares fit before evaluating it at $m = q$ using

$$\mathcal{D}(z) = \boldsymbol{\phi}(q)^{\mathrm{T}}\boldsymbol{\mathcal{B}}\boldsymbol{\mathcal{A}}^{\mathrm{T}}\boldsymbol{\mathcal{A}}\boldsymbol{\mathcal{F}}(z) \quad \text{(A.4)}$$

where, for $B = 2$,

$$\boldsymbol{\mathcal{B}} = \begin{bmatrix} 0 & -1 & 0 \\ 0 & 0 & -2 \\ 0 & 0 & 0 \end{bmatrix}. \quad \text{(A.5)}$$

The discrete-time transfer function $\mathcal{D}(z)$, may be realized using an IIR digital filter to compute the temporal derivatives required in gradient-based optical flow calculations, such as the Lucas-Kanade method used in [34]; however, this filter was not used in this paper. Like the prediction filter derived from $\mathcal{H}(z)$, the derivative filter may be tuned to attenuate high-frequencies, which is a useful property when the input is known to be corrupted by sensor noise. See [8] or [32] for an alternative derivation of the filters considered in this Appendix.

*Appendix B*

Like the Fourier spectrum, computed using the Discrete Fourier Transform (DFT), the Laguerre spectrum, computed using the Discrete Laguerre Transform (DLT) [33], may be used to form a compact representation of a signal and exploited for data compression or information extraction purposes. A finite window, possibly in conjunction with a tapering function, is used to isolate data prior to the application of the (short-time) DFT; however an exponentially decaying window of infinite extent is used to 'focus' the DLT. The DFT is best suited to signals that consist of wide-band sinusoidal components; whereas the DLT is more appropriate when the signals have a polynomial structure, or are comprised of very low-frequency sinusoids. Multidimensional Fourier-spectrum analysis techniques have a long history, especially in image/video analysis [18],[19],[31],[35]-[37], and are well established; however, the potential benefits of multidimensional Laguerre-spectrum exploitation are not so widely appreciated; therefore a technique, that is particularly useful in the current application, namely the estimation of point-target velocity, is presented in this Appendix.

In the spatial coordinates, the point target is assumed to be a scaled unit impulse, convolved with a 2-D Gaussian point-spread function (PSF). The target intensity $I_{\mathrm{tgt}}$, in the immediate vicinity of the maximum at $\boldsymbol{m} = [0 \quad 0 \quad 0]$ is modelled as a quadratic function using a linear combination of polynomial components

$$I_{\mathrm{tgt}}(m_x, m_y) = I_{\max} + \rho_x m_x^2 + \rho_y m_y^2 \quad \text{(B.1)}$$

where $I_{\max}$ is the maximum target intensity (typically positive) and $\rho$ is a PSF curvature parameter (typically negative). Now for a target moving with velocity $\boldsymbol{v}_{\mathrm{tgt}} = [v_x, v_y]$

$$I_{\mathrm{tgt}}(m_x, m_y, m_z) = I_{\max} + \rho_x(m_x - v_x m_z)^2 + \rho_y(m_y - v_y m_z)^2$$

$$= I_{\max} + \rho_x m_x^2 + \rho_y m_y^2 - 2\rho_x v_x m_x m_z - 2\rho_y v_y m_y m_z + \left(\rho_x v_x^2 + \rho_y v_y^2\right) m_z^2. \quad \text{(B.2)}$$

Equating the target parameters with polynomial component coefficients $\gamma_{k_x k_y k_z}$, and omitting the $m_z^2$ term, yields

$$\gamma_{000} = I_{\max}, \gamma_{200} = \rho_x, \gamma_{020} = \rho_y, \ \gamma_{101} = -2\rho_x v_x, \gamma_{011} = -2\rho_y v_y. \quad \text{(B.3)}$$

If estimates of the polynomial component coefficients $\hat{\gamma}$, are available, then the velocity estimate $\hat{\boldsymbol{v}}_{\text{tgt}}$, is simply computed from (B.3) using

$$\hat{v}_x = -\frac{\hat{\gamma}_{101}}{2\hat{\gamma}_{200}} \text{ and } \hat{v}_y = -\frac{\hat{\gamma}_{011}}{2\hat{\gamma}_{020}}. \quad \text{(B.4)}$$

(Note the similarity of Eq. B.4 and the motion constraint equations, used in Lucas-Kanade optical flow [34].)

Unfortunately, the analysis filters described in Section 2 yield the *Laguerre* polynomial coefficients $\hat{\beta}$, not the required polynomial *component* coefficients $\hat{\gamma}$, therefore a $\hat{\beta} \to \hat{\gamma}$ transformation step is required, involving the $\boldsymbol{\mathcal{A}}$ matrix from Appendix A, in the spatial and temporal dimensions. If $\hat{\boldsymbol{\gamma}}$ is a tensor of order three (corresponding to the $x$, $y$ and $z$ axes) with dimensions $\left(B_x + 1, B_y + 1, B_z + 1\right)$ and elements $\hat{\gamma}_{k_x k_y k_z}$, then for $B = 2$,

$$\hat{\boldsymbol{\gamma}} = \sum_{k_z=0}^{2} \sum_{k_y=0}^{2} \sum_{k_x=0}^{2} \hat{\beta}_{k_x k_y k_z} \left\{\boldsymbol{\mathcal{A}}_x^{\mathrm{T}} \boldsymbol{u}_{k_x}\right\} \otimes \left\{\boldsymbol{\mathcal{A}}_y^{\mathrm{T}} \boldsymbol{u}_{k_y}\right\} \otimes \left\{\boldsymbol{\mathcal{A}}_z^{\mathrm{T}} \boldsymbol{u}_{k_z}\right\} \quad \text{(B.5)}$$

where $\otimes$ is a tensor product (or vector outer product) and $\boldsymbol{u}_k = [\delta_{0,k} \quad \delta_{1,k} \quad \delta_{2,k}]^{\mathrm{T}}$ (i.e. unit basis vectors in the Laguerre space). If only estimates of the $\{\gamma_{000}, \gamma_{200}, \gamma_{020}, \gamma_{101}, \gamma_{011}\}$ coefficients are required to model the moving point target, as suggested above in (B.1)-(B.3), then the structure of the $\boldsymbol{\mathcal{A}}$ matrices dictates that only the $\{\beta_{000}, \beta_{010}, \beta_{020}, \beta_{100}, \beta_{200}, \beta_{011}, \beta_{101}\}$ coefficients need to be estimated during analysis processing in stage 2 and considered when accumulating the target power using (18b).

When modeling the foreground target signal in this way, two important points should be kept in mind when choosing the decay rate of the 3-D exponential weighing function $w(\boldsymbol{m})$. On the one hand, a *rapidly decaying* function strongly emphasizes the region close to $\boldsymbol{m} = [0 \quad 0 \quad 0]$, therefore the ability to faithfully represent rapid motion is limited. On the other hand, the modeled target intensity may become very negative before the weight becomes negligible, if a *gradually decaying* function is used. These opposing considerations unfortunately reduce the fidelity of this approach for fast targets; however the computational savings brought about by recursion may compensate for the loss of accuracy in real-time applications.

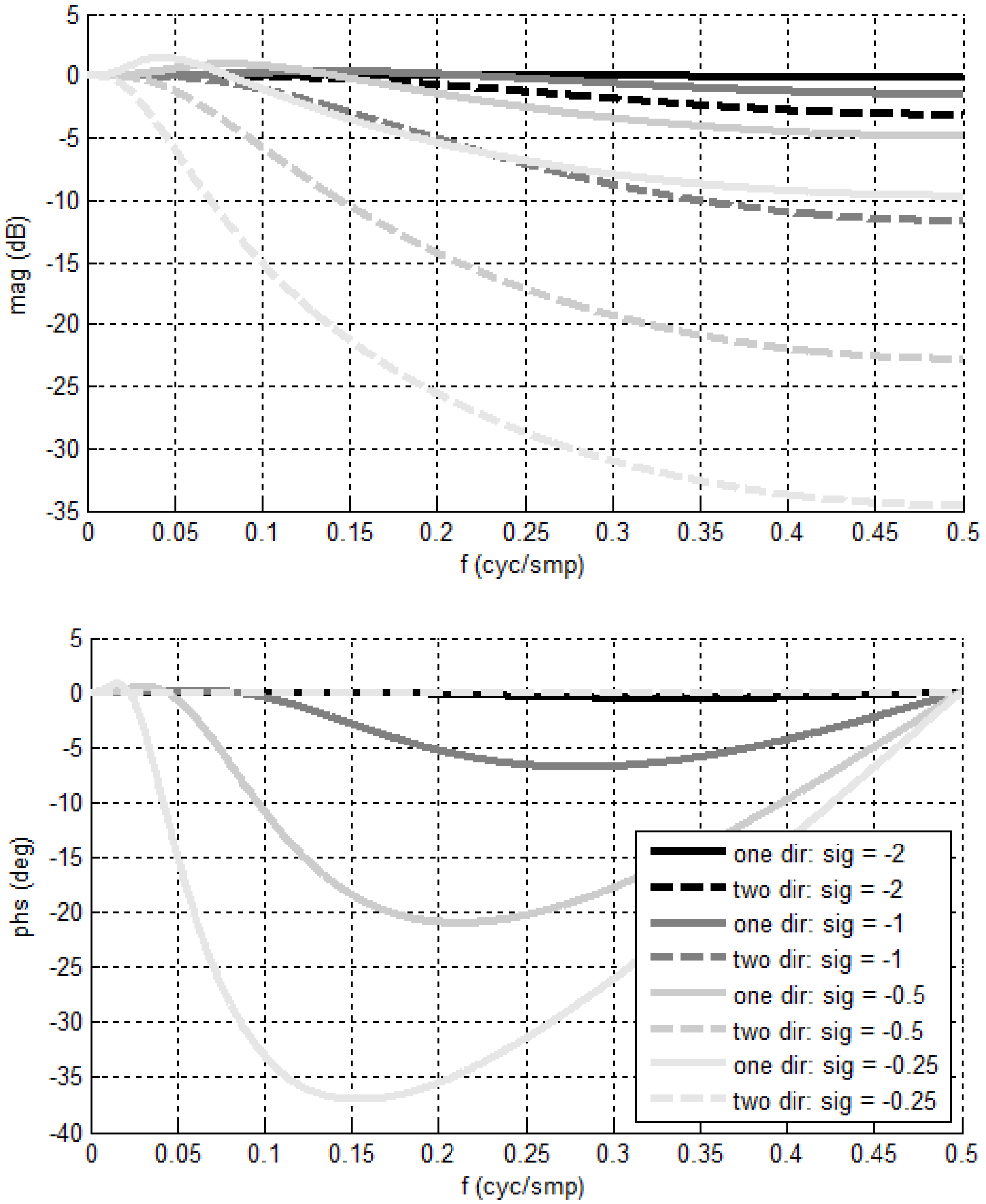

**Fig. 1**. Magnitude and phase responses, as a function of normalized frequency (cycles per sample) for causal (one dir) and non-causal (two dir) filters with $q = 0$, and various $\sigma$ (sig) values. Note that the phase is zero for all non-causal filters.

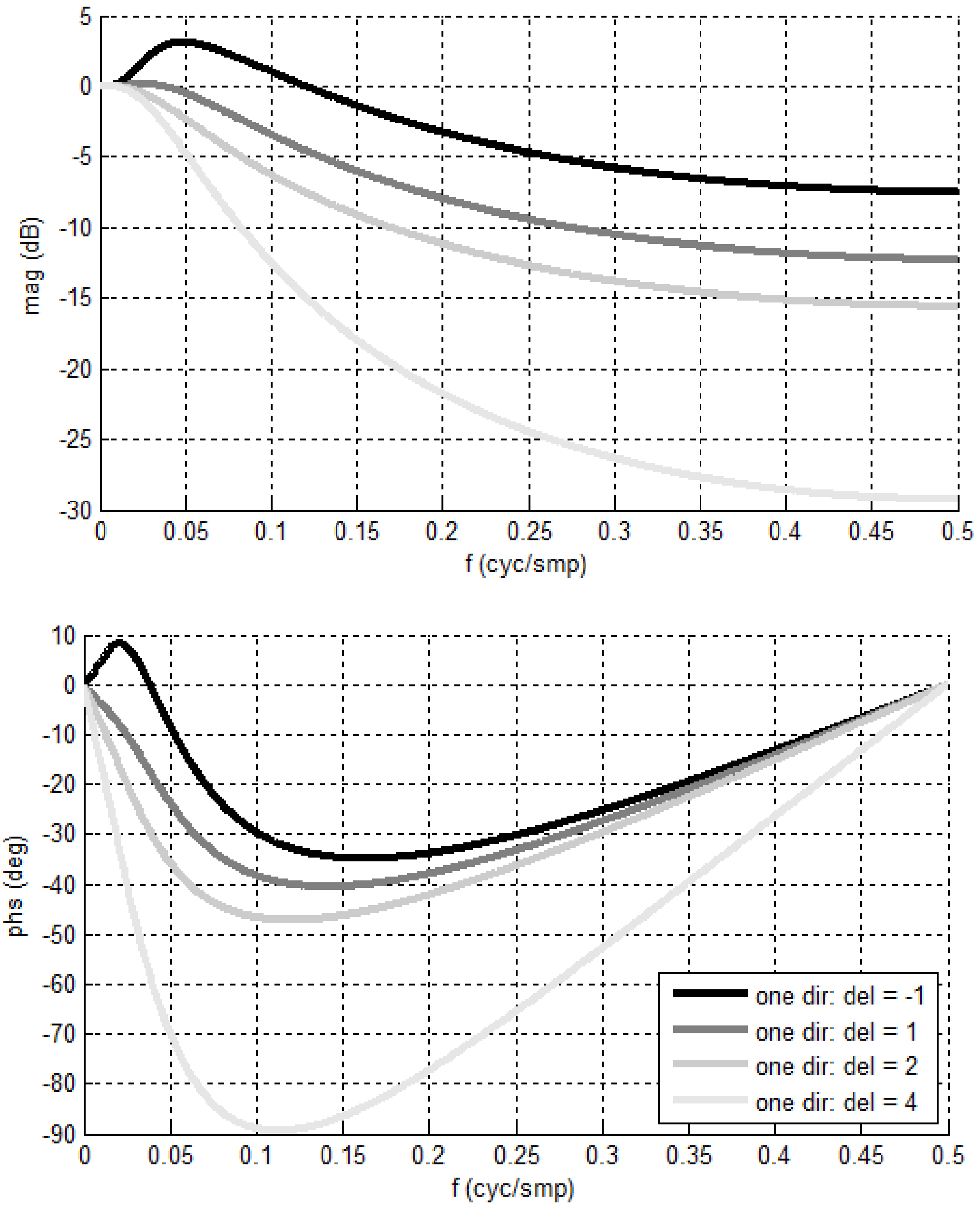

**Fig. 2**. Frequency responses for causal (one dir) filters with $\sigma = -1/4$ and various $q$ (del) values.

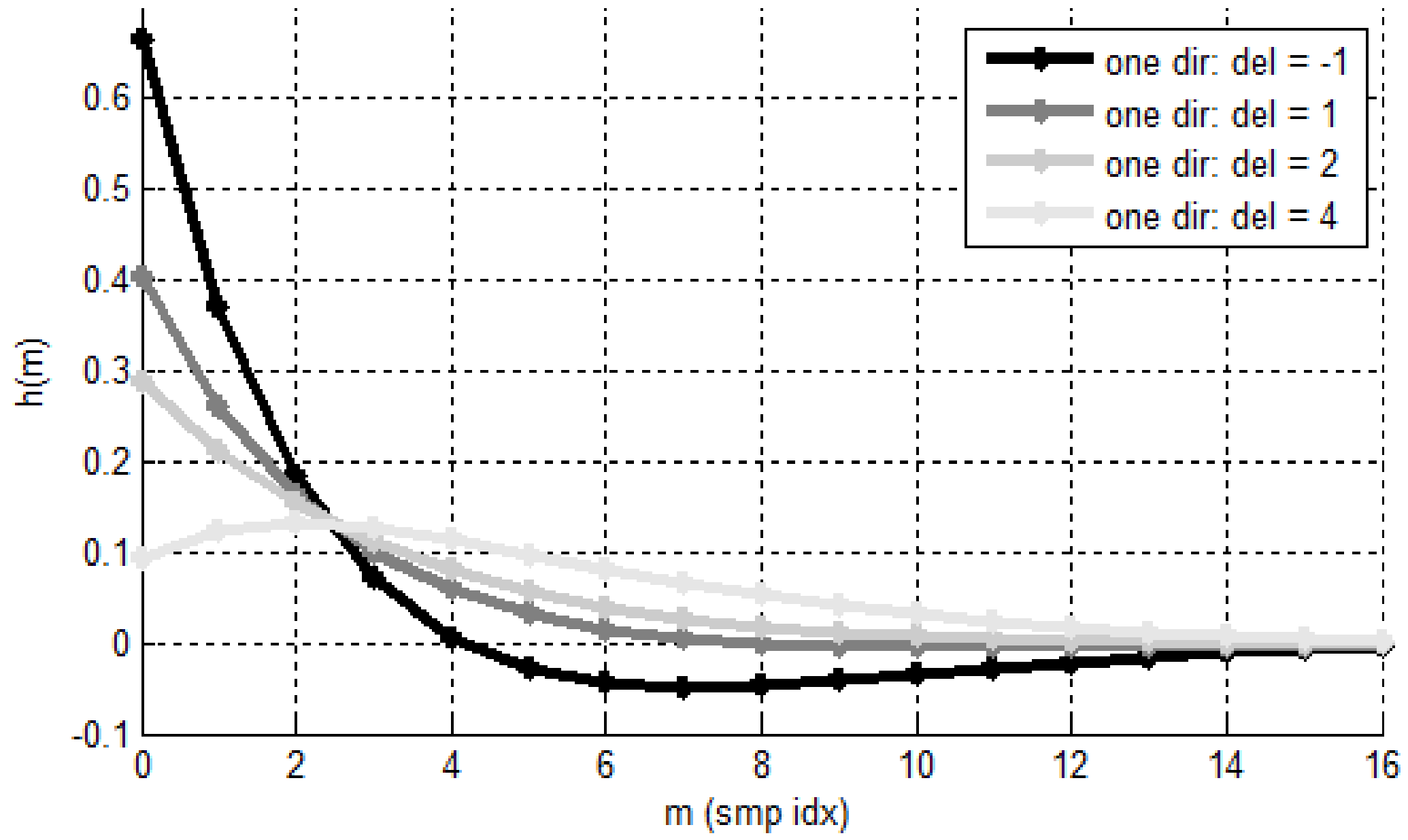

**Fig. 3**. Impulse responses for causal (one dir) filters with $\sigma = -1/4$ and various $q$ (del) values.

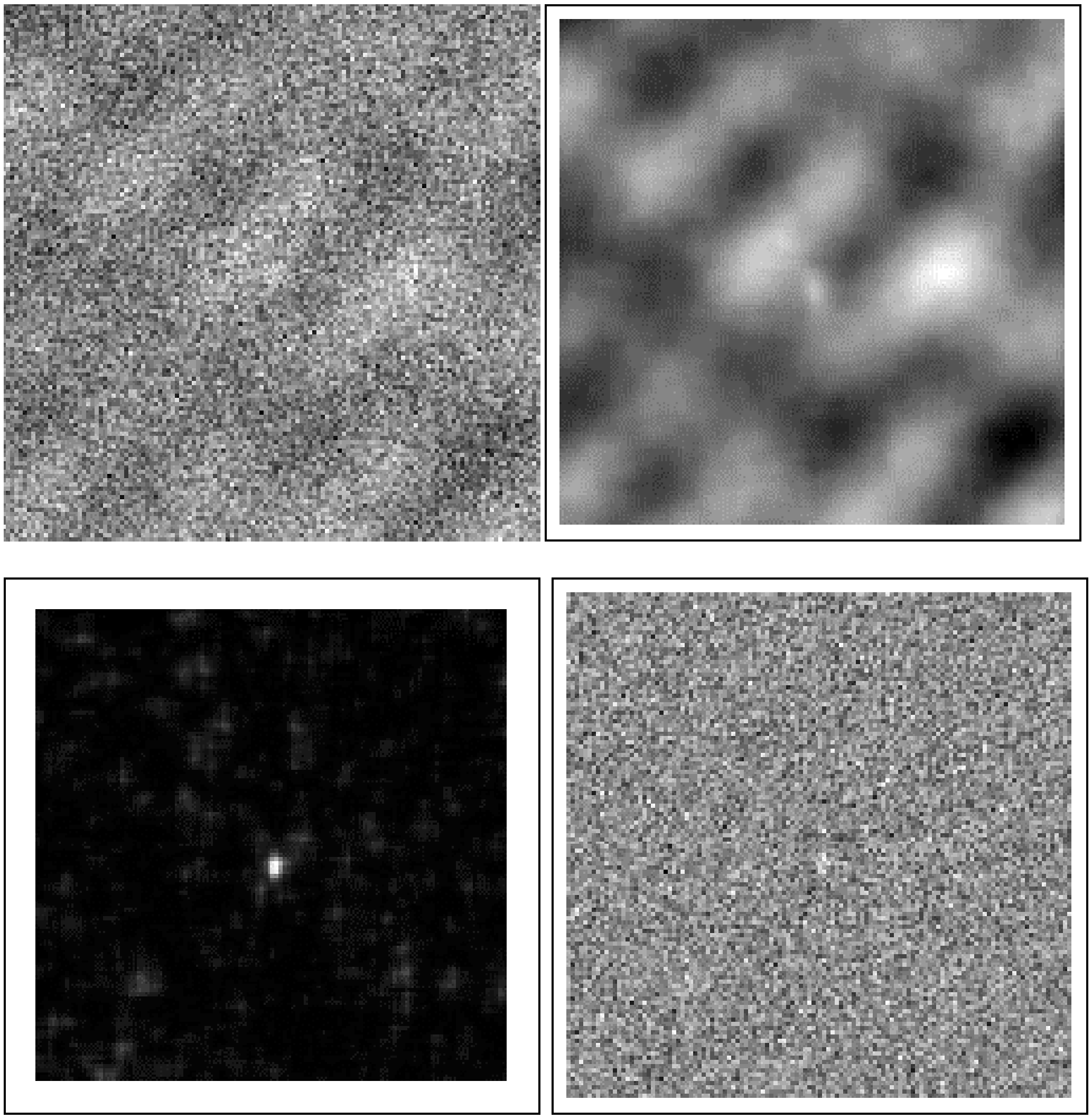

**Fig. 4**. Processing-sequence example, processed images cropped to remove (spatial) filter startup transients. Clockwise from top left: stage 1 input, raw image, $J$; estimated background, $\hat{I}$; stage 1 output, stage 2 input, background subtracted, $I_{\epsilon}$; stage 2 ouput, accumulated foreground power, $\hat{P}_{\epsilon}$, target clearly visible at centre of FOV.

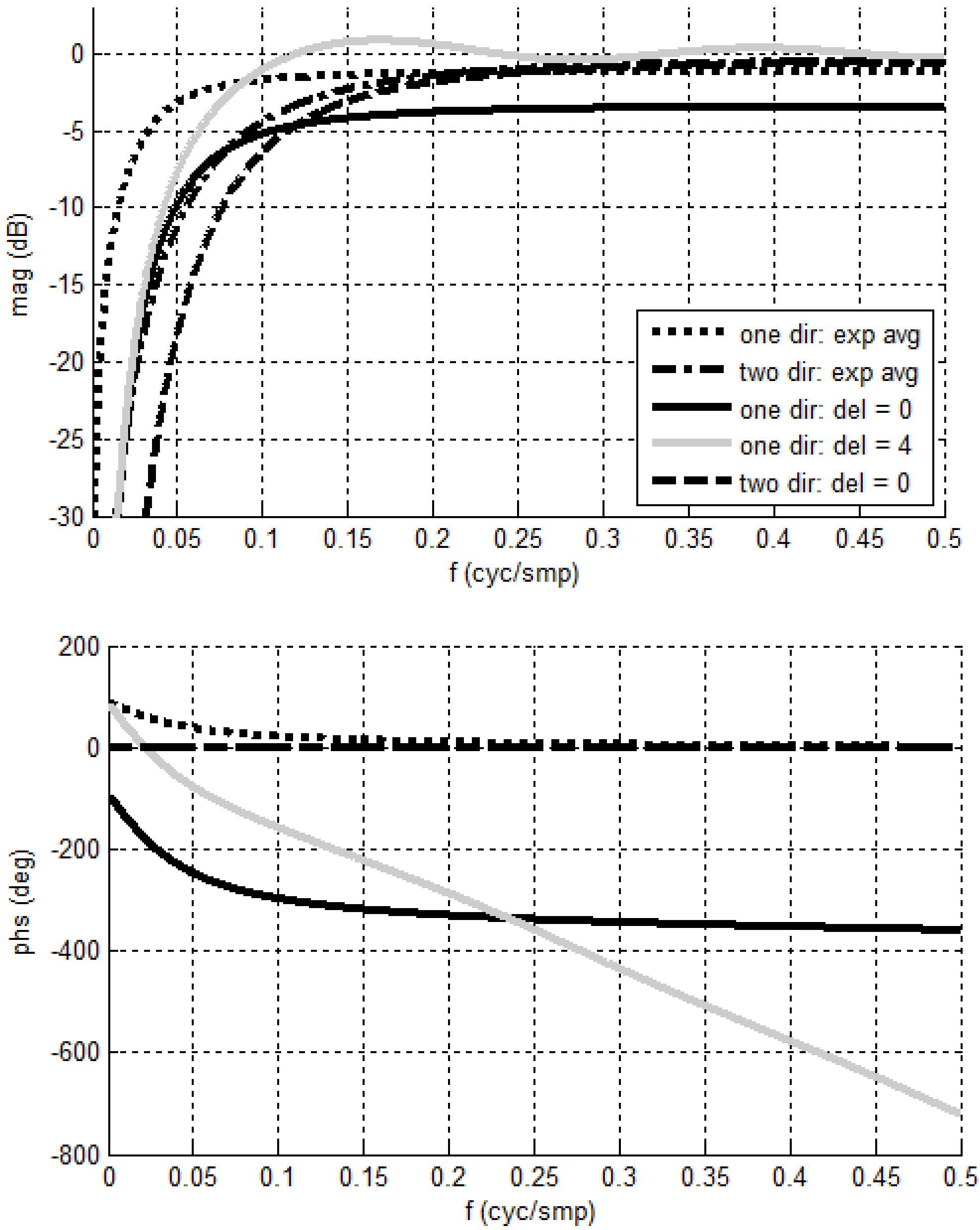

**Fig. 5**. Frequency responses for the high-pass filters used in stage 1 ($B = 2$). Non-causal (two dir) filter with $\sigma = -1/2$ and causal (one dir) filters with $\sigma = -1/4$ and two different $q$ (del) values. Causal and non-causal exponential average subtraction filters ($B = 0$) also shown for comparison.

TABLE IA: STAGE 1 FILTER COEFFICIENTS, CAUSAL

| | $m$ | 0 | 1 | 2 | 3 |
|---|---|---|---|---|---|
| Low-pass | $b$ | 0.0920 | -0.0913 | 0.0102 | 0 |
| | $a$ | 1.0000 | -2.3364 | 1.8196 | -0.4724 |

TABLE IB: STAGE 1 FILTER COEFFICIENTS, NON-CAUSAL

| | $m$ | 0 | 1 | 2 | 3 |
|---|---|---|---|---|---|
| Low-pass | $b$ | 0.1463 | -0.0925 | -0.0561 | 0.0327 |
| Fwd & Bwd | $a$ | 1.0000 | -1.8196 | 1.1036 | -0.2231 |

TABLE IIA: STAGE 2 FILTER COEFFICIENTS, CAUSAL

| | $m$ | 0 | 1 | 2 | 3 |
|---|---|---|---|---|---|
| $k = 0$ | $b$ | 0.4703 | 0 | 0 | 0 |
| | $a$ | 1.0000 | -0.7788 | 0 | 0 |
| $k = 1$ | $b$ | -0.4151 | 0.4151 | 0 | 0 |
| | $a$ | 1.0000 | -1.5576 | 0.6065 | 0 |
| $k = 2$ | $b$ | 0.3663 | -0.7326 | 0.3663 | 0 |
| | $a$ | 1.0000 | -2.3364 | 1.8196 | -0.4724 |

TABLE IIB: STAGE 2 FILTER COEFFICIENTS, NON-CAUSAL

| | $m$ | 0 | 1 | 2 | 3 |
|---|---|---|---|---|---|
| $k = 0$ | $b$ | 0.2474 | 0.1501 | 0 | 0 |
| Fwd & Bwd | $a$ | 1.0000 | -0.6065 | 0 | 0 |
| $k = 1$ | $b$ | 0 | +/-0.1072 | 0 | 0 |
| Fwd / Bwd | $a$ | 1.0000 | -1.2131 | 0.3679 | 0 |
| $k = 2$ | $b$ | -0.1093 | 0.0832 | 0.0505 | -0.0244 |
| Fwd & Bwd | $a$ | 1.0000 | -1.8196 | 1.1036 | -0.2231 |

TABLE III: LINEAR DIFFERENCE EQUATION (LDE) COEFFICIENTS*

| |
|---|
| **Stage 1 (background estimator) filter coefficients [a]:** |
| *Causal, Low-pass:*<br>$c = (1-p)/2$<br>$b_0 = c(q^2p^2 + 3qp^2 + 2p^2 - 2q^2p + 2p + q^2 - 3q + 2)$<br>$b_1 = -c(2q^2p^2 + 8qp^2 + 6p^2 - 4q^2p - 4qp + 6p + 2q^2 - 4q)$<br>$b_2 = c(q^2p^2 + 5qp^2 + 6p^2 - 2q^2p - 4qp + q^2 - q)$, $b_3 = 0$<br>$a_0 = 1,\ a_1 = -3p, a_2 = 3p^2, a_3 = -p^3$ |
| *Non-causal (Fwd & Bwd) Low-pass:*<br>$c = 2(p^2 + 8p + 1)$<br>$b_0 = \frac{1}{c}(p^2 + 10p + 1)(1-p)/(1+p)$<br>$b_1 = \frac{3}{c}p(p^2 - 1)$, $b_2 = \frac{3}{c}p^2(p^2 - 1)$<br>$b_3 = \frac{1}{c}p^3(p^2 + 10p + 1)(1-p)/(1+p)$<br>$a_0 = 1,\ a_1 = -3p, a_2 = 3p^2, a_3 = -p^3$ |
| **Stage 2 (foreground accumulator) filter coefficients [b]:** |
| *Causal,* $k = 0$:<br>$b_0 = \sqrt{1-p}, b_1 = 0$<br>$a_0 = 1,\ a_1 = -p$ |
| *Causal,* $k = 1$:<br>$c = -\sqrt{p(1-p)^3}/(1-p)$<br>$b_0 = c,\ \ b_1 = -c, b_2 = 0$<br>$a_0 = 1,\ \ a_1 = -2p,\ \ a_2 = p^2$ |
| *Causal,* $k = 2$:<br>$c = p\sqrt{(1-p)^5}/(1-p)^2$<br>$b_0 = c, b_1 = -2c, b_2 = c, b_3 = 0$<br>$a_0 = 1,\ a_1 = -3p, a_2 = 3p^2, a_3 = -p^3$ |
| *Non-Causal (Fwd & Bwd),* $k = 0$<br>$c = \frac{1}{2}\sqrt{(1-p)/(1+p)}$<br>$b_0 = c, b_1 = cp$<br>$a_0 = 1,\ a_1 = -p$ |
| *Non-Causal (Fwd / Bwd),* $k = 1$<br>$b_0 = 0, b_1 = (+/-)\frac{1}{2}\sqrt{2p(1-p)^3/(1+p)}$, $b_2 = 0$<br>$a_0 = 1,\ a_1 = -2p,\ a_2 = p^2$ |
| *Non-Causal (Fwd & Bwd),* $k = 2$<br>$c = \sqrt{2}(1-p)^2\sqrt{p^3 + 9p^2 + 9p + 1}$<br>$b_0 = \frac{-1}{c}\sqrt{p(1-p)^5}$, $b_1 = \frac{1}{c}\sqrt{p(1-p)^5}(p^2 - p + 1)$<br>$b_2 = \frac{1}{c}\sqrt{p^3(1-p)^5}(p^2 - p + 1)$, $b_3 = \frac{-1}{c}\sqrt{p^7(1-p)^5}$<br>$a_0 = 1,\ a_1 = -3p, a_2 = 3p^2, a_3 = -p^3$ |

*Pole location: $p = e^{\sigma}$, $B = 2$.

[a] Analysis and synthesis; real poles with a multiplicity of $B + 1$; nominal delay of $q$ samples applied on synthesis.

[b] Analysis-only; real poles with a multiplicity of $k + 1$.

TABLE A.I: DISCRETE LAGUERRE POLYNOMIAL COEFFICIENTS ($\alpha_{k\acute{k}}$), WHERE $p = e^{\sigma}$

| | $\acute{k} = 0$ | $\acute{k} = 1$ | $\acute{k} = 2$ |
|---|---|---|---|
| $k = 0$ | $\sqrt{1-p}$ | 0 | 0 |
| $k = 1$ | $-\sqrt{p(1-p)}$ | $\sqrt{(1-p)^3/p}$ | 0 |
| $k = 2$ | $p\sqrt{(1-p)^5}/(1-p)^2$ | $-(3p+1)\sqrt{(1-p)^5}/[2p(1-p)]$ | $\sqrt{(1-p)^5}/(2p)$ |